\documentclass[runningheads]{llncs}

\usepackage{float}
\usepackage{graphicx}
\usepackage{amssymb}
\usepackage{dsfont}
\usepackage{amstext}
\usepackage{amsmath}
\DeclareMathOperator*{\argmax}{arg\,max}

\begin{document}
\title{DIAR: Deep Image Alignment and Reconstruction using Swin Transformers}
%
%
\author{Monika Kwiatkowski\inst{1}\orcidID{0000-0001-9808-1133} \and
Simon Matern\inst{1}\orcidID{0000-0003-3301-2203} \and
Olaf Hellwich\inst{1}\orcidID{0000-0002-2871-9266}}
%
%
\institute{Computer Vision \& Remote Sensing, Technische Universität Berlin, Marchstr. 23, Berlin, Germany}
\maketitle              
\begin{abstract}
When taking images of some occluded content, one is often faced with the problem that every individual image frame contains unwanted artifacts, but a collection of images contains all relevant information if properly aligned and aggregated. In this paper, we attempt to build a deep learning pipeline that simultaneously aligns a sequence of distorted images and reconstructs them. 
We create a dataset that contains images with image distortions, such as lighting, specularities, shadows, and occlusion. We create perspective distortions with corresponding ground-truth homographies as labels. We use our dataset to train Swin transformer models to analyze sequential image data. The attention maps enable the model to detect relevant image content and differentiate it from outliers and artifacts. We further explore using neural feature maps as alternatives to classical key point detectors. The feature maps of trained convolutional layers provide dense image descriptors that can be used to find point correspondences between images. We utilize this to compute coarse image alignments and explore its limitations.

\end{abstract}
\newpage
\section{Introduction}

This paper attempts to solve the problem of image reconstruction and alignment simultaneously. Specifically, we deal with image sets that contain distortions and are related by a 2D homography. When taking photos of panoramas or planar objects, such as magazines, paintings, facades, or documents, the resulting images often contain unwanted artifacts. The images may contain varying lighting conditions, shadows, and occlusions. Each artifact corrupts the original content. In order to combine information from all images, the images have to be aligned, and the information has to be aggregated.
In this work, we utilize deep image features for image alignment. Furthermore, we use Video Swin Transformers for spatio-temporal analysis of the aligned image sequences and aggregation. This paper is a continuation of the work by Kwiatkowski and Hellwich (2022) 
\cite{kwiatkowski2022specularity}. 

We provide the following improvements to the initial paper:
\begin{enumerate}
    \item We developed a new synthetic dataset. We improve the original data generation by using a ray-tracing pipeline. The dataset contains more realistic lighting and shadows. Furthermore, we take images from different perspectives and provide ground-truth homographies. The library can further be used for image alignment tasks.
    \item In addition to solving an image reconstruction task, we also align the images using their neural feature maps. We use feature maps as dense key point descriptors and compute matches between images using a cosine similarity score. 
    \item Following the success of Vision Transformers (ViT) \cite{dosovitskiy2020image,liu2021swin} and their extensions to video \cite{arnab2021vivit,liu2022video} we explore their use for image reconstruction. (Video) Vision transformers compute spatio-temporal attention maps. They have been shown to be able to compute spatial features that are on par with Convolutional neural networks. Furthermore, transformers are the state of the art deep learning models for sequential data. Both of these properties are essential for image reconstruction. A local image feature can only be determined as an artifact by analyzing the context within the image and across the image sequence.
    \item We explore various forms of aggregating feature maps. We show that computing attention maps over the sequence allows for a better aggregation compared to the original concept of Deep Sets.
\end{enumerate}

\section{Related Work}
\subsubsection{Deep Image Alignment}
The classical approach to estimating homographies uses sparse image descriptors to compute matches between image points \cite{hartley2003multiple}. A homography can then be estimated from the matches using random sample consensus (RANSAC). A variety of methods have been developed to improve on the classical approach with neural networks.
Some methods aim at replacing image descriptors, such as SIFT\cite{lowe2004distinctive}, with trainable feature maps \cite{hpatches_2017_cvpr,lindenberger2021pixel,shen2020ransac}. The advantage of these methods is that they can learn robust image features that are distinctive in their corresponding dataset. Additionally, they can be easily integrated into many classical computer vision pipelines.\\
Neural networks have also been trained on matching image patches directly \cite{Sun_2021_CVPR,sarlin2020superglue}. Graph neural networks or transformers enable a model to analyze points not only individually but in relation to each other. A computed match should therefore be consistent with neighboring matches; otherwise, it is likely an outlier. \\
One can also describe the estimation of a homography as a regression problem \cite{detone2016deep}. The model takes two images as input and outputs eight values that can be interpreted as the parameters of a homography.

\subsubsection{Deep Image Stitching} combines information from multiple images. The images are simultaneously aligned and stitched along overlapping regions. This way, image alignment and image reconstruction can be combined into a single differentiable end-to-end pipeline \cite{nie2021unsupervised}. There have also been attempts to use neural representations of images for image alignment and stitching. BARF \cite{lin2021barf} uses neural radiance fields to store 2D or 3D scenes inside a neural network. During the training of the network, both the content of the scene and the relative orientation of cameras are estimated. BARF uses a form of Bundle Adjustment that produces very accurate results. However, none of these methods can deal with outliers and artifacts. 
Furthermore, BARF does not learn any prior that is transferable to other image sequences. 

\subsubsection{(Vision) Transformers} have become the state-of-the-art model for sequential data \cite{vaswani2017attention,devlin2018bert}. Transformers compute attention maps over the whole input sequence. Each token in a sequence is compared with every other token, which enables embeddings with a global context. It has been shown that Transformers can also be applied to image tasks \cite{dosovitskiy2020image,liu2021swin}. Images can be treated as a sequence of image patches. The same idea has also been also extended to video \cite{arnab2021vivit,liu2022video}. Videos can be split into spatio-temporal patches and processed as a sequence. In this work, we want to train video transformers to aggregate image sequences. Using spatial and temporal information, it should be possible to determine whether a patch contains a defect or an outlier. The reconstruction should then reconstruct the underlying content by combining partial information from the larger context of the sequence.

\section{Dataset}

\begin{figure}[H]
    \centering
    \includegraphics[width=0.9\linewidth]{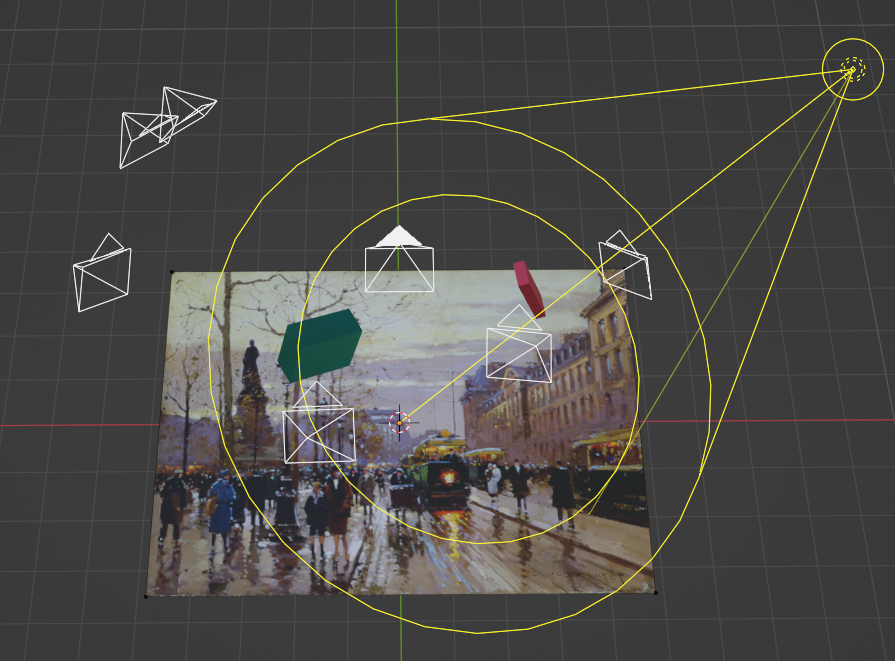}
    \caption{Illustration of a randomly generated scene using Blender. The plane shows a painting. The white pyramids describe randomly generated cameras; the yellow cone describes a spotlight. Geometric objects serve as occlusions and cast shadows onto the plane.}
    \label{fig:blender}
\end{figure}

Figure \ref{fig:blender} illustrates the data generation process. We use paintings from the Wiki Art dataset \cite{danielczuk2019segmenting} as our ground-truth labels for reconstruction. Any other image dataset could also be used, but Wiki Art contains a large variety of artworks from various periods and art styles. We believe the diversity of paintings makes the reconstruction more challenging and reduces biases towards a specific type of image. We take an image from the dataset and use it as a texture on a plane in 3D. 
We randomize the generation of cameras by sampling from a range of 3D positions. The field of view of the camera is also randomly sampled to create varying zoom effects. This also creates a variety of intrinsic camera parameters. We randomly generate light sources. The type of light source is randomly sampled, e.g., spotlight, point light, or area light, their position, and corresponding parameters, such as intensity or size. \\
Furthermore, we generate geometric objects and put them approximately in between the plane and the camera's positions. We utilize Blender's ability to apply different materials to textures. We apply randomized materials to the image texture and occluding objects. The appearance of occluding object can be diffuse, shiny, reflective, and transparent. The material properties also change the effect lighting has on the plane.
It changes the appearance of specularities, shadows, and overall brightness. \\
Finally, we iterate over the cameras and render the images. Blender's physically-based path tracer, Cycles, is used for rendering the final image. Path-tracing enables more realistic effects compared to rasterization. It allows the simulation of effects, such as reflections, retractions, and soft shadows, which are not possible in the original dataset generation pipeline \cite{kwiatkowski2022specularity}. \\
Using this data generation, we create two datasets. One contains misaligned images; the other contains aligned images.
\subsection{Aligned Dataset}
\label{seq:aligned}
We create one dataset that only consists of aligned image sequences. To enforce the alignment, we use a single static camera that perfectly fits the image plane. The camera's viewing direction is set to be perpendicular to the image plane and centered on the image plane. We adjust the vertical and horizontal field of view such that only the image can be seen. Furthermore, we generate a ground truth label by removing all light sources and occluding objects. We only use ambient illumination for rendering the picture.

\begin{figure}[H]
    \centering  
    \includegraphics[width=0.49\linewidth]{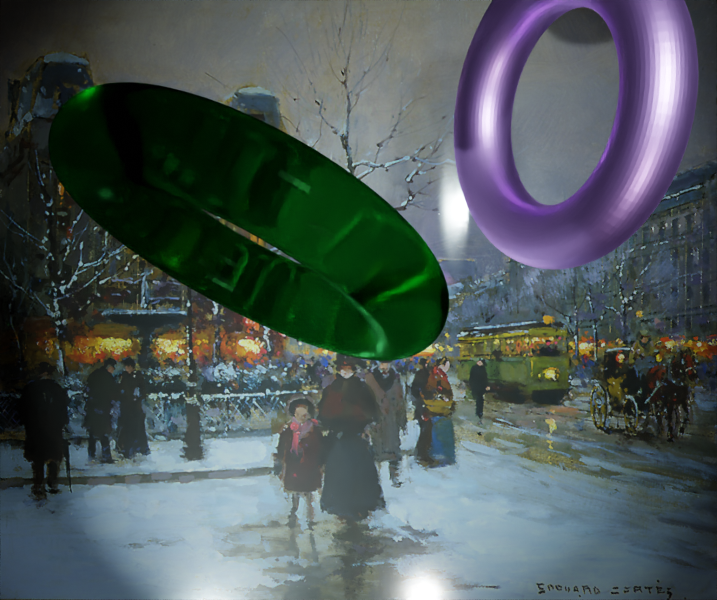}
    \includegraphics[width=0.49\linewidth]{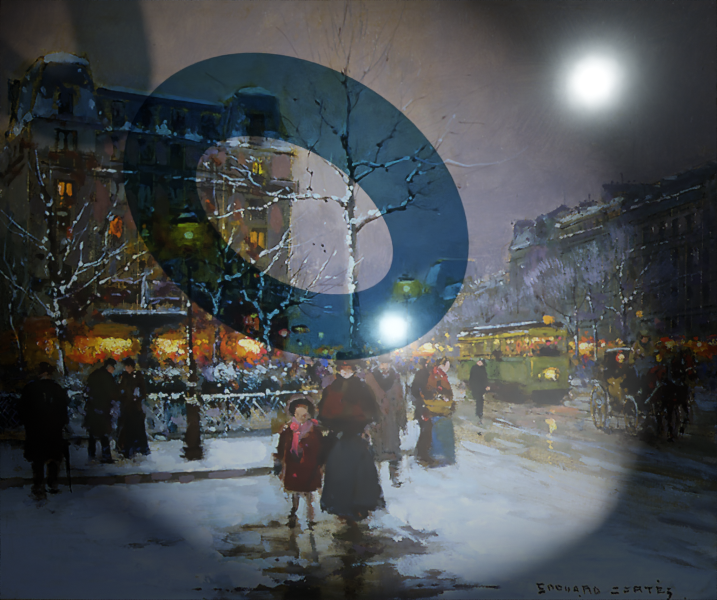}
    \includegraphics[width=0.49\linewidth]{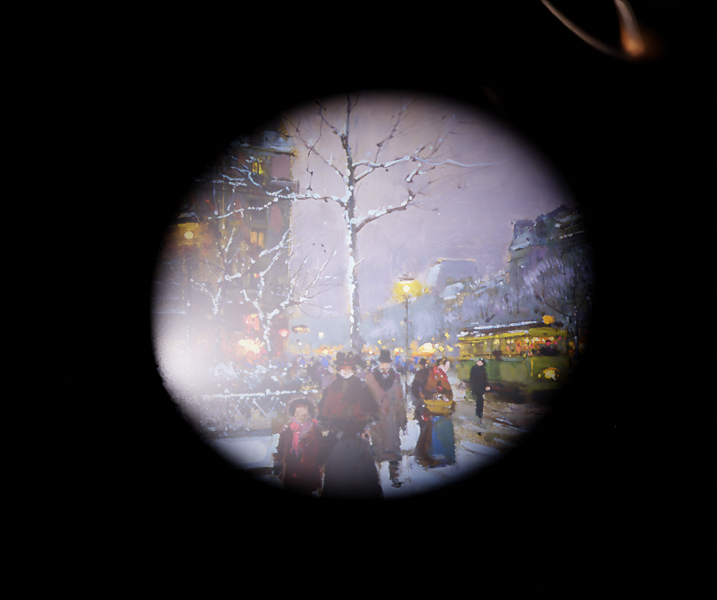}
    \includegraphics[width=0.49\linewidth]{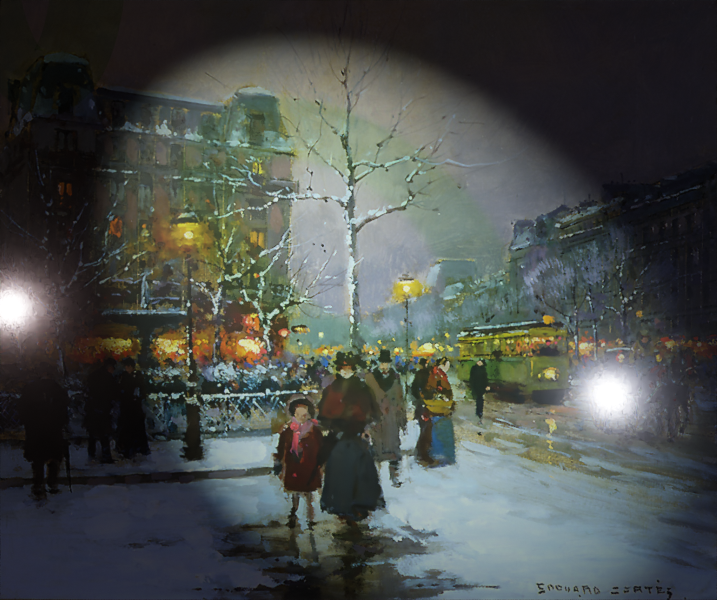}
    \caption{Four randomly generated images that are aligned.}
    \label{fig:aligned}
\end{figure}

\begin{figure}[H]
    \centering
    \includegraphics[width=0.7\linewidth]{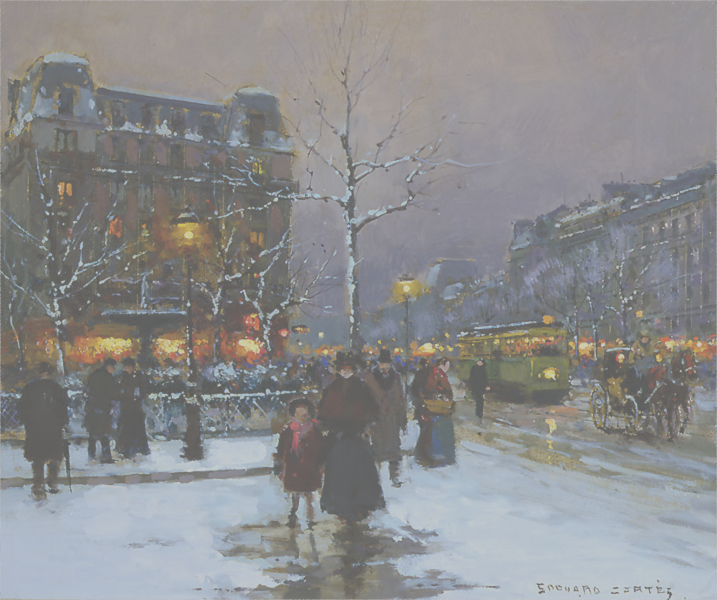}
    \caption{The image shows the output of our rendering pipeline when only ambient lighting is used. The image is free of any artifacts.}
    \label{fig:aligned_gt}
\end{figure}
Figure \ref{fig:aligned} shows a sequence of distorted images that are also aligned. Figure \ref{fig:aligned_gt} shows the corresponding ground-truth label. We generate $\sim 15000$ image sequences, each containing ten distorted images as our dataset. This dataset is used for training the image reconstruction task.
We create another test set with 100 sequences, each containing 50 distorted images. We use the test set to analyze how well our models perform on varying sequence sizes.
\subsection{Misaligned Dataset}
\label{sec:misaligned}
\begin{figure}[H]
    \centering  
    \includegraphics[width=0.49\linewidth]{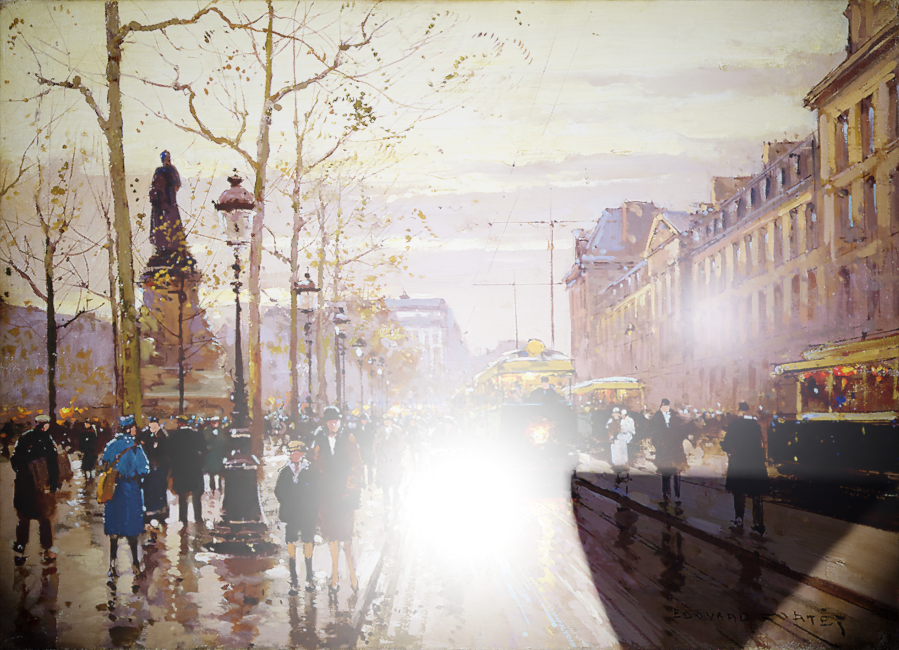}
    \includegraphics[width=0.49\linewidth]{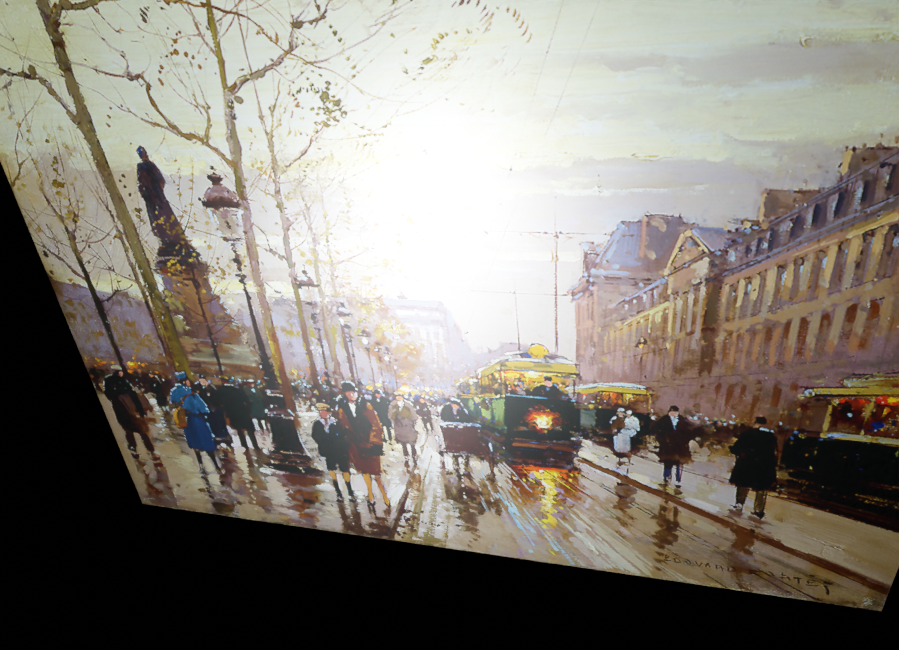}
    \includegraphics[width=0.49\linewidth]{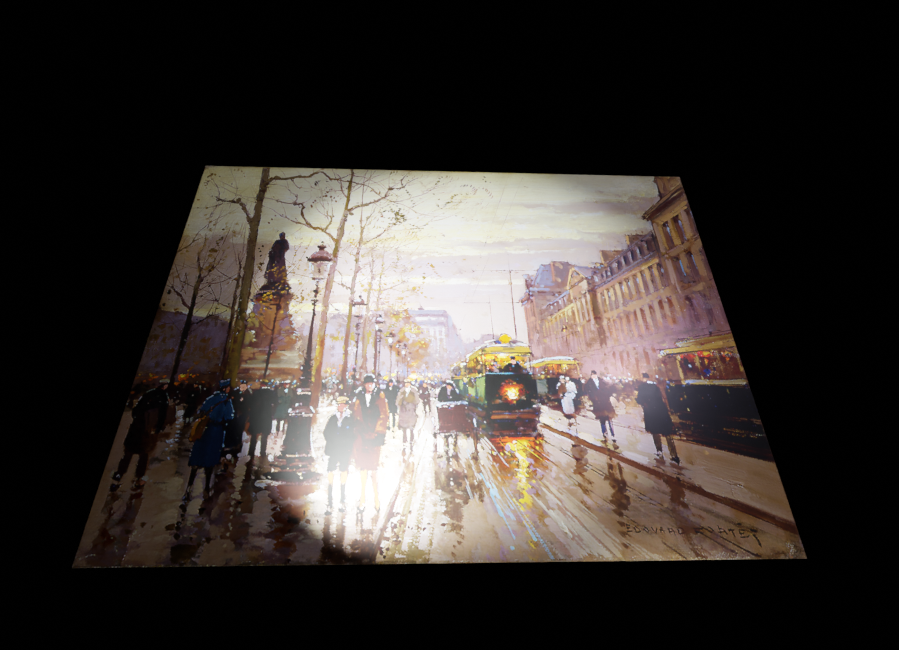}
    \includegraphics[width=0.49\linewidth]{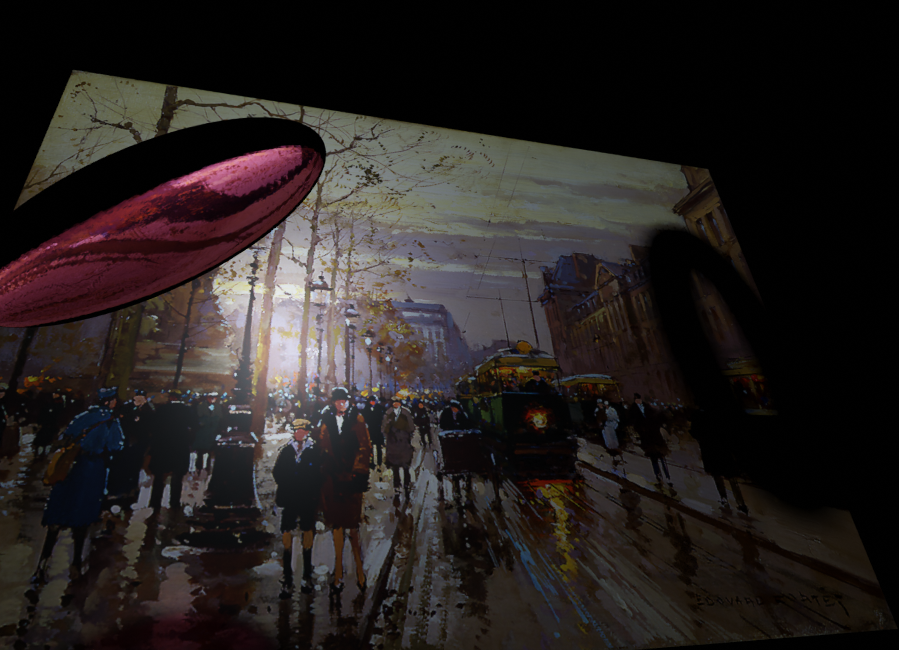}
    \caption{Four images containing perspective distortions. The first image is aligned with the camera's view, but it also contains image distortions.}
    \label{fig:misaligned}
\end{figure}

In addition to the aligned dataset, we generate images with perspective distortions. For each randomly generated camera, we render an image. In order to evaluate the image alignment with reconstruction, we also generate a single ground truth image as described in section \ref{seq:aligned} under ambient lighting conditions. We need a common reference frame that is aligned with the ground-truth label in order to measure alignment and reconstruction simultaneously. Using the label directly as input creates an unwanted bias for the model. Therefore, another aligned image is created that contains distortions. \\
Figure \ref{fig:misaligned} shows a sequence of distorted images. The first image contains various artifacts, but it is free of perspective distortions. 

\subsubsection{Homography}
\label{sec:homography}
Since we take images of planar objects, all our images are related by 2D homographies. A point with pixel coordinates $(x_i,y_i)$ in image $I_i$ is projected onto the coordinates $(x_j,y_j)$ in image $I_j$ with the homography $H_{ij}$:
\begin{align}
    \lambda \begin{pmatrix}
    x_j \\ y_j \\ 1
    \end{pmatrix} = \underbrace{    \begin{bmatrix}
        h_{11} &h_{12} & h_{13}\\
        h_{11} &h_{22} & h_{23}\\
        h_{31} &h_{32} & h_{33}
    \end{bmatrix}}_{H_{ij}}
    \begin{pmatrix}
    x_i \\ y_i \\ 1
    \end{pmatrix}
\end{align}
$H_{ij}$ can be exactly computed from four point-correspondences. Since we know the camera's positions in space and their intrinsic parameters, we can calculate the projection points explicitly. We project each of the four corners $X_k ~~k=1,..,4$ of the paintings into the $i$-th camera image plane $x_k^{(i)}$ using their projection matrices:
\begin{align}
    x_k^{(i)} &= P X_k \\
   &= K_i [I|0]  \begin{bmatrix}
    R_i & t_i\\ 0 & 1
    \end{bmatrix} X_k ~~~~k=1,..,4
\end{align}
$(R_i, t_i)$ describes the global rotation and translation of the camera, $K_i$ describes the intrinsic camera parameters. Using the four-point pairs $x_k^{(i)} \leftrightarrow x_k^{(j)} ~~k=1,..,4$, we can compute the homography using the Direct Linear Transform (DLT)\cite{hartley2003multiple}. 
\section{Deep Image Alignment}

\begin{figure}[H]
    \centering
    \includegraphics[width=0.7\linewidth]{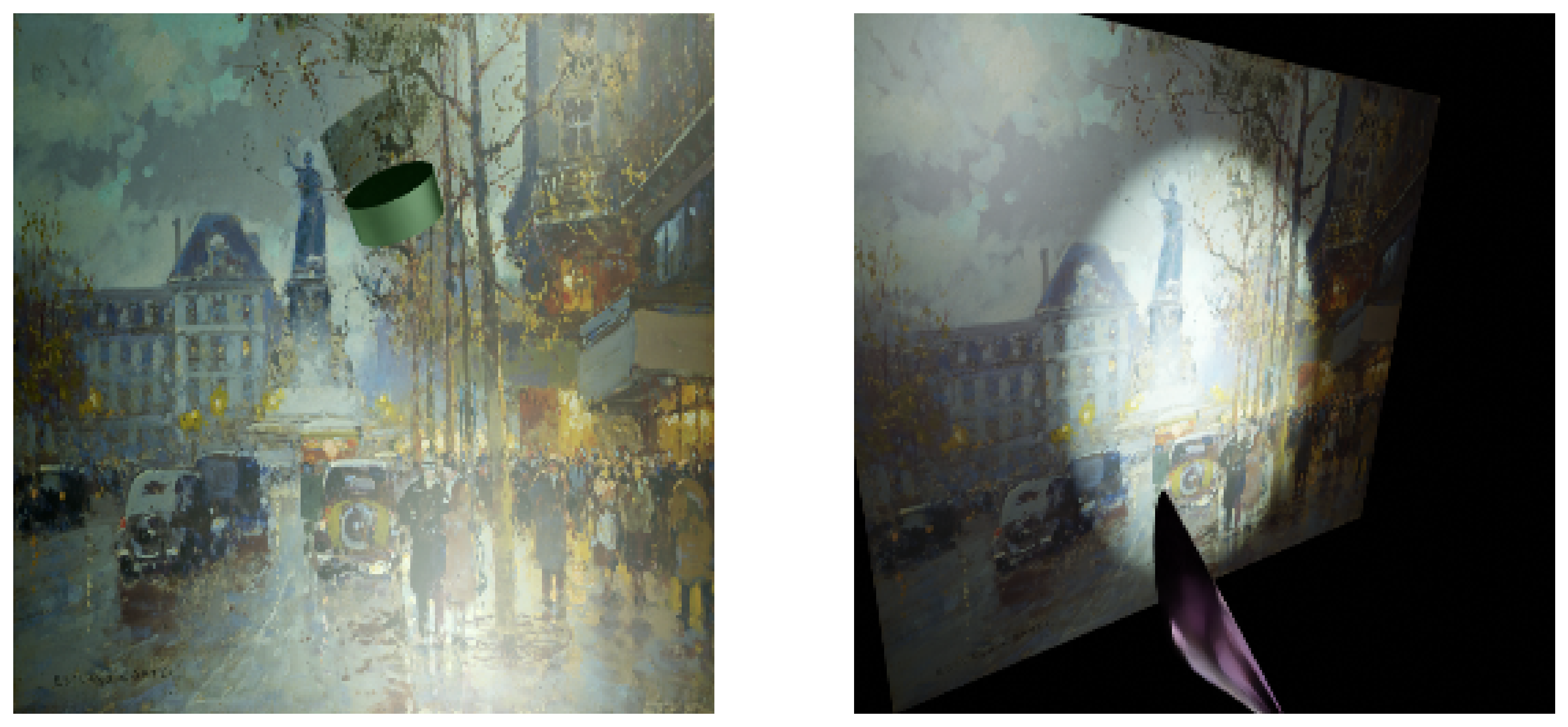}
    \includegraphics[width=0.7\linewidth]{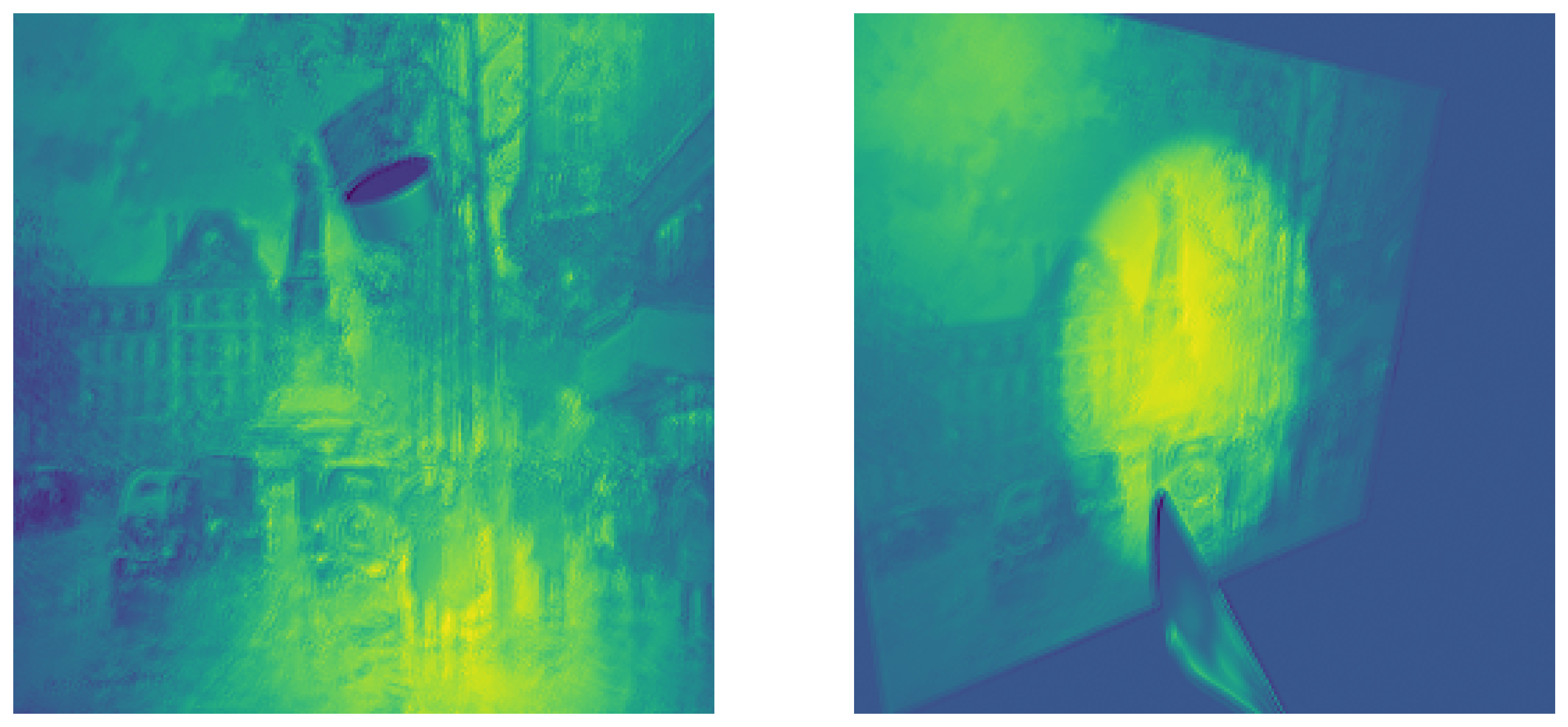}
    \caption{A pair of images and their corresponding feature maps. }
    \label{fig:feature}
\end{figure}

In order to align the images, we chose to use convolutional feature maps as image descriptors. It has been shown by methods, such as Ransac-Flow \cite{shen2020ransac} or Pixel-Perfect SfM\cite{lindenberger2021pixel}, that it is possible to use convolutional layers as dense keypoint descriptors. Figure \ref{fig:feature} shows two images and their corresponding feature maps.

Let $I\in \mathbb{R}^{H \times W}$ be an image and $I'=f_\theta(I)\in \mathbb{R}^{H \times W  \times C}$ a feature map from a convolutional neural network. 
We can treat each individual pixel in $I'$ as a key point with a descriptor of dimension $C$
Using this approach any image of resolution $(H\times W)$ can be described as a collection of points described by a data matrix $X_I=(x_{1},\cdots,x_{WH})^T \in \mathbb{R}^{HW  \times C}$. Given two collection of keypoints $X_{1}$ and $X_{2}$ we can compute similarity scores $S_{ij}$ between any point $x_i\in X_{1}$ and $x_j\in X_{2}$. We use the cosine similarity:
\begin{align}
    S(x_i,x_j) &= \frac{<x_i,x_j>}{|x_i|\cdot |x_j|} \\
    S &= \text{norm}(X_{2}X_{1}^T)
\end{align}

The score matrix $S$ can be computed and does not require the images to have the same resolution. Every pixel in one image is compared with every pixel in another image. 
We can compute matches by filtering pairs with a maximal similarity score. Each point pair should be each other's best match; otherwise, there is ambiguity between descriptors. If $x_i = \argmax_k S(x_k,x_j)$ and $x_j = \argmax_k S(x_i,x_k)$, then $(x_i,x_j)$ are a match. \\
The score matrix $S$ and their matches can be computed efficiently. We additionally follow the implementation of Ransac-Flow by using image pyramids to make the descriptors and matching more scale invariant. For any image at scale $I_s \in \mathbb{R}^{sH \times sW}$, we can again compute keypoints $X_{I_s} \in\mathbb{R}^{k^2HW  \times C}$.
We can concatenate the descriptors into a single matrix $X \in\mathbb{R}^{N  \times C} $, where $N=\sum_k k^2HW$. We can compute matches as before. \\
We can use the matches to estimate our homographies as described in \ref{sec:misaligned}. Using RANSAC, we can further filter out matches and estimate a homography. Given a homography described as a function $H_{ij}:(x,y)\mapsto (x',y')$, we can warp image $I_i(x,y)$ into $I_j(x,y)$ using $I_i(H_{ij}(x,y))$. \\
This is a general approach that can be integrated into any differentiable pipeline. For our implementation, we use the first three layers of a ResNet that was pre-trained on ImageNet1K.

\section{Architecture}

\subsection{Deep Residual Sets}

For our architecture, we use Deep Residual Sets as our baseline. Deep sets can be decomposed into an encoder $\phi$ and a decoder $\rho$ as follows\cite{zaheer2017deep}:
\begin{align}
    f(x_1,x_2,\cdots, x_N) = \rho \left( \sum_{i=1}^N\phi(x_i)  \right)
\end{align}

Deep sets have useful properties, such as permutation invariance. In the original approach, Deep Residual Sets have shown promising results for image reconstruction\cite{kwiatkowski2022specularity}. The architecture consists of residual blocks, downsampling, and upsampling layers. Further average pooling was used to aggregate the embeddings along the sequence. 

\begin{figure}[H]
    \centering
    \includegraphics[width=\linewidth]{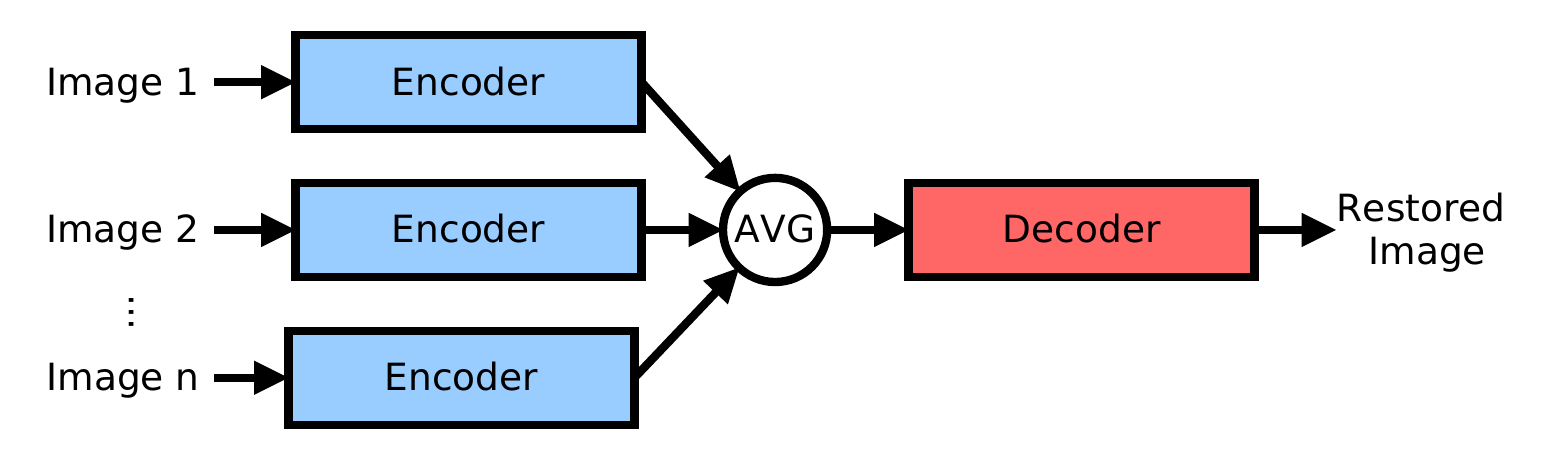}
    \caption{Initial architecture (figure taken from\cite{kwiatkowski2022specularity})}
    \label{fig:my_label}
\end{figure}

The main disadvantage of deep sets is that they struggle to remove outliers along the sequence. If an image contains an occlusion, the corresponding embedding of the outlier is pooled into the final embedding. The decoder often struggles to remove them and can only attenuate these artifacts. \\ 
In order to remove outliers, they must first be identified within the sequence using contextual information. Deep sets analyze each element independently and, therefore, can't solve this problem. Transformer models, on the other hand, provide a better solution for this problem since they compute attention maps over the whole sequence. 

\subsection{Video Swin Transformer}

\begin{figure}[H]
    \centering
    \includegraphics[width=\linewidth]{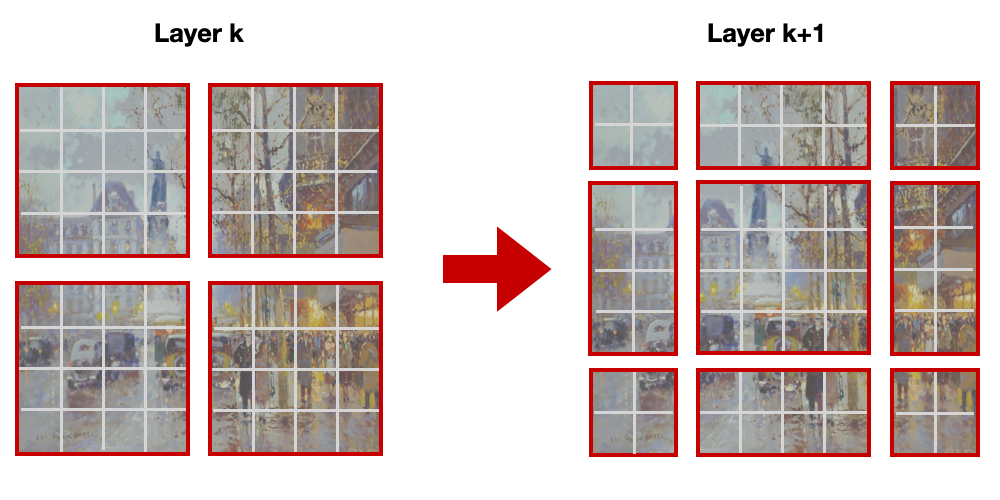}
    \caption{An illustration of the partitioning scheme of two consecutive Swin layers. The input of layer $k$ consists of $W\times W$ pixels. Using a window size of $M\times M$ creates $\frac{W}{M}\times \frac{W}{M}$ patches. In layer $k+1$ the windows are shifted by $\frac{M}{2}\times \frac{M}{2}$}
    \label{fig:swin}
\end{figure}

Transformers have been shown to be powerful sequential models \cite{devlin2018bert,vaswani2017attention}. Transformers compute attention over a sequence by computing pairwise embeddings between tokens. This principle allows for very general sequence processing. However, their main disadvantage is their quadratic memory consumption with regard to the sequence length. Applying them to high-dimensional images is not a trivial task. Transformers have been successfully applied to vision tasks by using special partitioning schemes on images\cite{dosovitskiy2020image}. Images aren't processed pixel by pixel but rather over larger patches. The original Vision Transformer (ViT) has a low inductive bias, which allows them to learn more general feature extraction, but also increases training time. Several architectures have been proposed to improve the efficiency of vision transformer models. \\
Swin transformers provide an efficient way to process images and videos as sequences \cite{liu2021swin}. Figure \ref{fig:swin} illustrates how a Swin transformer operates. Given is an image consisting of $8\times8$ pixels. Using a predefined window size, here $4\times4$, the $k$-th layer splits the input into $2\times 2$ patches. The patches are passed to a transformer layer, and self-attention maps are computed.
In the next layer, the windows are shifted by half the window size using a cyclic shift. The new patches are also passed to a self-attention layer. The combination of both layers allows efficient computation of attention across non-overlapping 

The same concept has been extended to video with Video Swin Transformers \cite{liu2022video}. The window size is extended with a time dimension. Given a video with size $T\times H \times W$ and a window size $P\times M \times M$ the video is divided into spatio-temporal patches of size $\frac{T}{P}\times \frac{H}{M} \times \frac{W}{M}$.

\subsection{Image Reconstruction using Swin Transformers}
\label{sec:swin}
\begin{figure}[H]
    \centering
    \includegraphics[width=\linewidth]{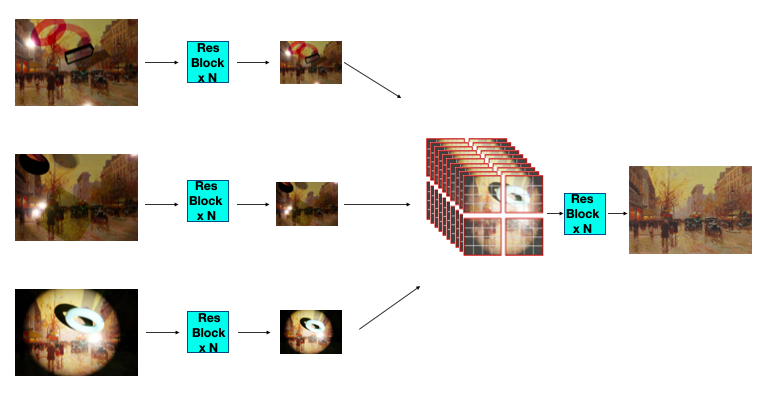}
    \caption{The architecture uses residual blocks for encoding and decoding. }
    \label{fig:swin-architecture}
\end{figure}

Although Deep Residual Sets provide a good baseline for image reconstruction, their main disadvantage is their lack of contextual information between images. We would like to alleviate this disadvantage by replacing the pooling layer with a transformer model. Specifically, we apply a Video Swin Transformer on the concatenated embeddings of the individual images. Figure \ref{fig:swin-architecture} illustrates our architecture. We use residual blocks for the encoding of each individual image and for the final decoding. The feature maps are concatenated, processed as a sequence, and given into the Swin layers. The Swin transformer is primarily used for aggregating information across the sequence. We use the downsampling and upsampling layers within our residual blocks. We do not use any downsampling or merging layers from the original Swin transformer. 

We explore three different Swin modules by changing the temporal window size. We use three models DIAR$(1,7,7)$, DIAR$(2,7,7)$ and DIAR$(3,7,7)$. Our graphics card could not handle larger patch sizes.

Although the Swin layer computes attention maps over the whole sequence, it does not aggregate information into a single element as a pooling layer does. 
Let $ [x_1,\cdots,x_T]\in \mathbb{R}^{(T,H,W,C)}$ be the stacked feature maps with dimensionality $(H,W,C)$. Let $[e_1,\cdots,e_T]=Swin(x)\in \mathbb{R}^{(T,H,W,C)}$ be the computed embedding from the Swin layer.
We explore various methods of aggregating the stacked embeddings into a single feature map for reconstruction.
\begin{itemize}
    \item Average pooling without embedding (Deep sets): $y = \sum_i x_i$ 
    \item Average pooling with embedding: $y = \sum_i e_i$
    \item Weighted sum: $y = \sum x_i\sigma(e)_i$ , where $\sigma( )$ describes the softmax function.
\end{itemize}

\subsection{Training}

All models are trained on the synthetic dataset of aligned sequences. The dataset contains 15000 image sequences. We use 10\% as a validation set.
We train our models on an NVIDIA RTX 3090 with 24 GB memory.
We train with a batch size of 20 for 100 epochs. We use the Adam optimizer with a learning rate of $\lambda= 0.001$. 

\section{Evaluation:}

\subsection{Aggregation}
\begin{figure}
    \centering
    \includegraphics[width=0.8\linewidth]{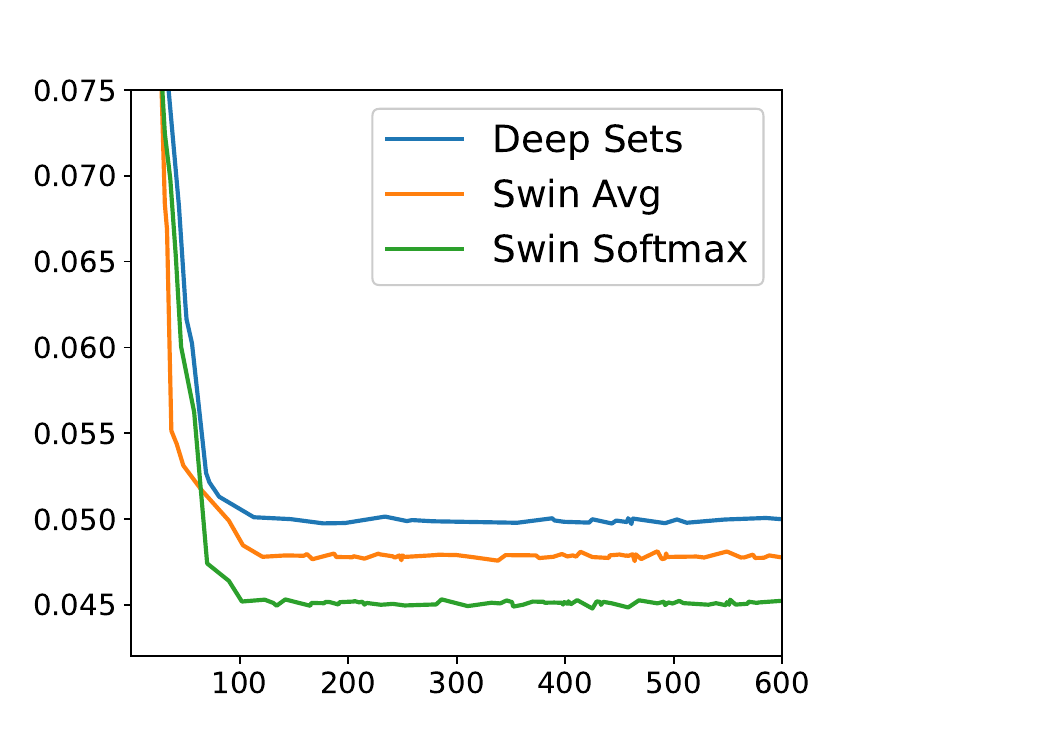}
    \caption{Deep sets clearly fall behind attention-based models. Aggregating the embeddings using a weighted sum provides the best reconstruction.}
    \label{fig:agg}
\end{figure}

As mentioned in section \ref{sec:swin} we evaluate different aggregation methods. We compare the original Deep Sets with Swin-based methods. Figure \ref{fig:agg} shows the progression of the models on the validation set performance.  

The graphic clearly shows that transformers improve aggregation. 
The figure also indicates that computing a softmax and aggregating the individual feature maps with a weighted sum is superior to average pooling over the Swin embedding. 
Based on these results, all further Swin models use softmax.

\newpage
\subsection{Image Reconstruction}

We use a test set containing 100 sequences, each with 50 distorted images. To evaluate our models, we use the metrics root-mean-squared error (RMSE), peak signal-to-noise ratio (PSNR), and structural similarity index measure (SSIM). In addition to our deep learning methods, we compare them with non-deep learning methods that provide a reasonable baseline for methods that deal with outlier removal. The following methods are also evaluated:
\begin{itemize}
    \item Median of images
    \item Average of images
    \item Robust PCA (RPCA)\cite{bouwmans2018applications},
    \item Intrinsic image decomposition (MLE)\cite{weiss2001deriving}
\end{itemize}

\begin{figure}[H]
    \centering
    \includegraphics[width=\linewidth]{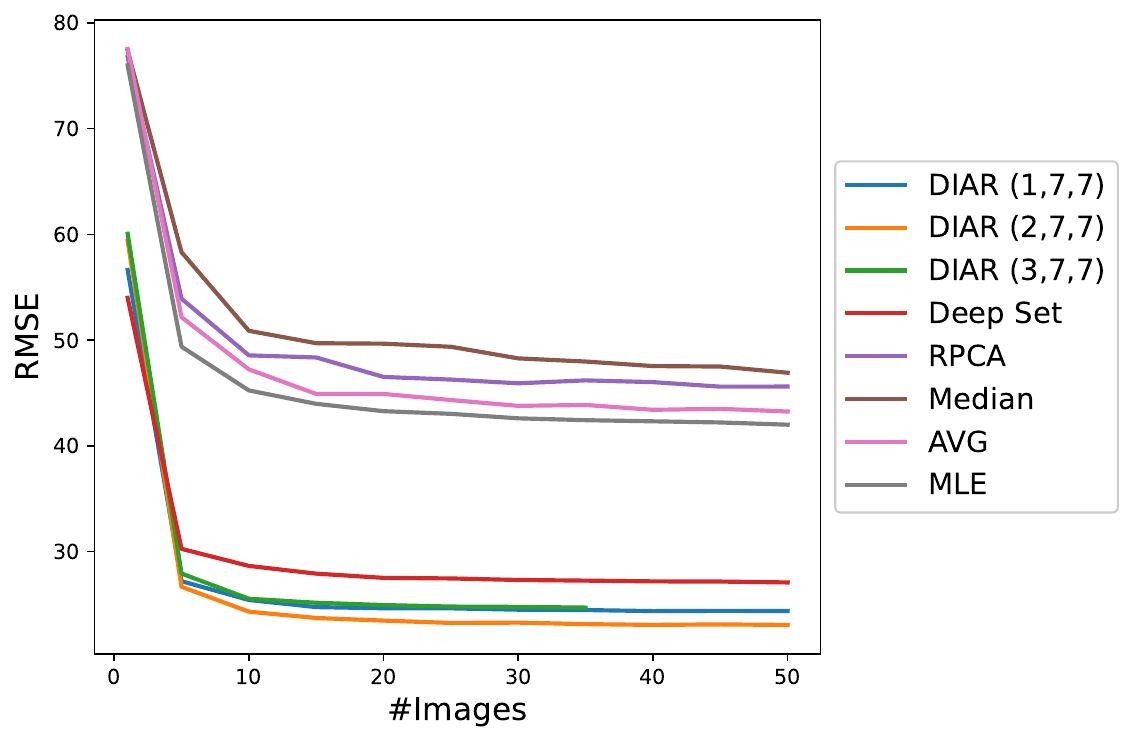}
    \caption{The graph shows the average RMSE for different input lengths}
    \label{fig:rmse}
\end{figure}

\begin{figure}[H]
    \centering
    \includegraphics[width=1\linewidth]{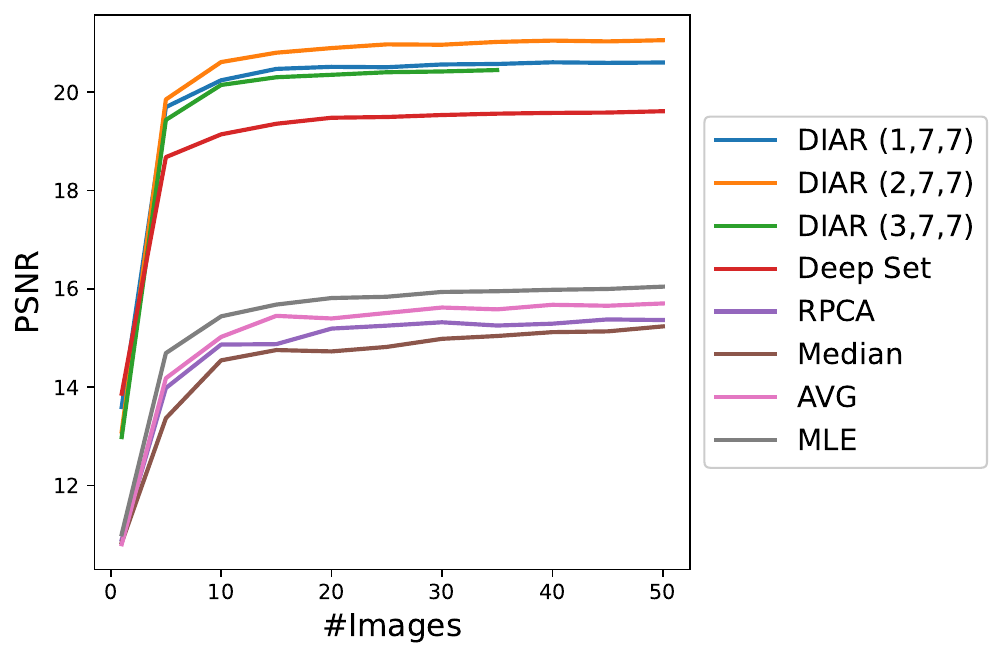}
    \caption{The graph shows the average PSNR for different input lengths}
    \label{fig:psnr}
\end{figure}

\begin{figure}[H]
    \centering
    \includegraphics[width=1\linewidth]{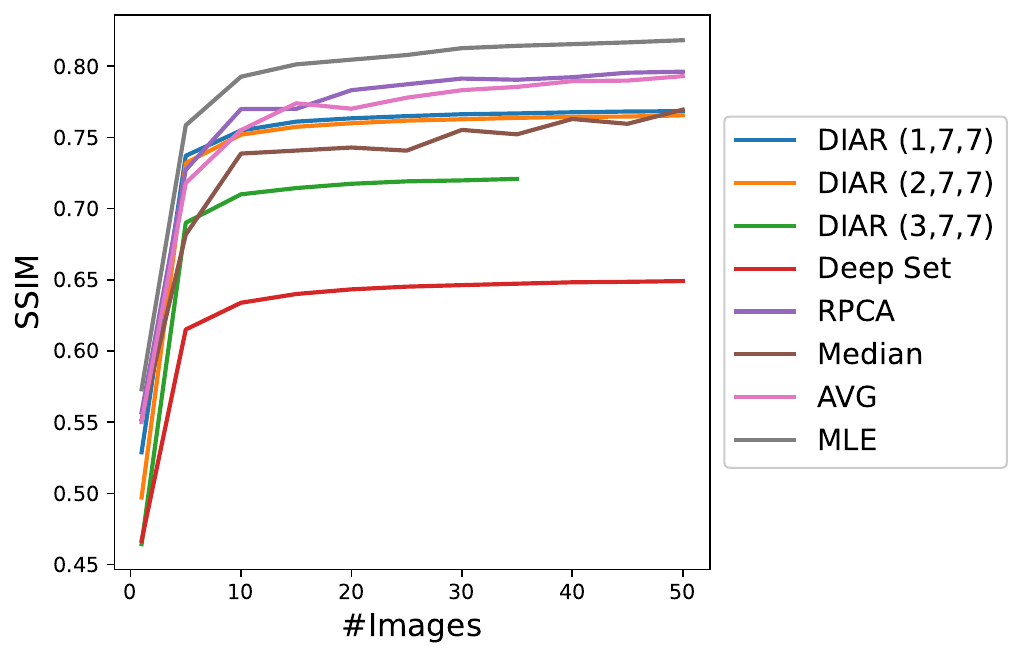}
    \caption{The graph shows the average SSIM for different input lengths}
    \label{fig:ssim}
\end{figure}

The figures \ref{fig:rmse},\ref{fig:psnr} and \ref{fig:ssim} illustrate the average performance of each model given different sequence lengths. The deep learning models have a by far lower RMSE and higher PSNR. All deep learning methods have low SSIM.

\begin{figure}[H]
    \centering
    \includegraphics[width=1\linewidth]{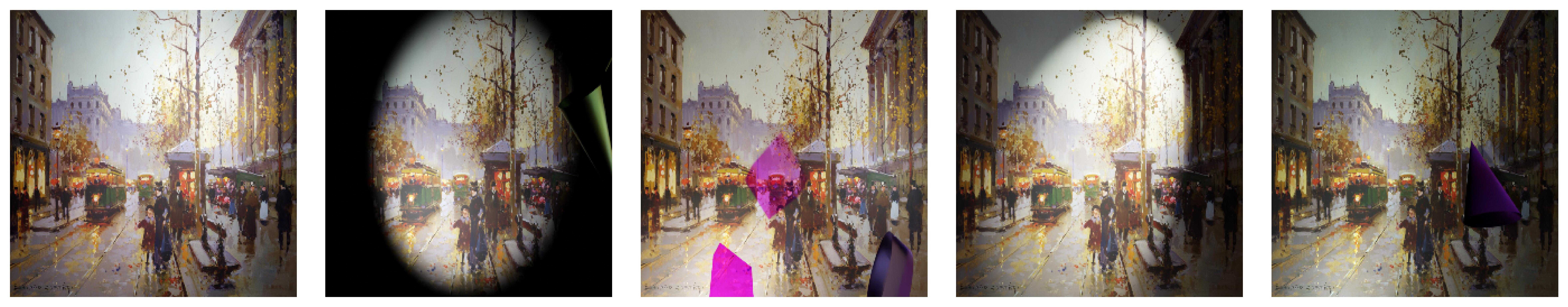}
    \includegraphics[width=1\linewidth]{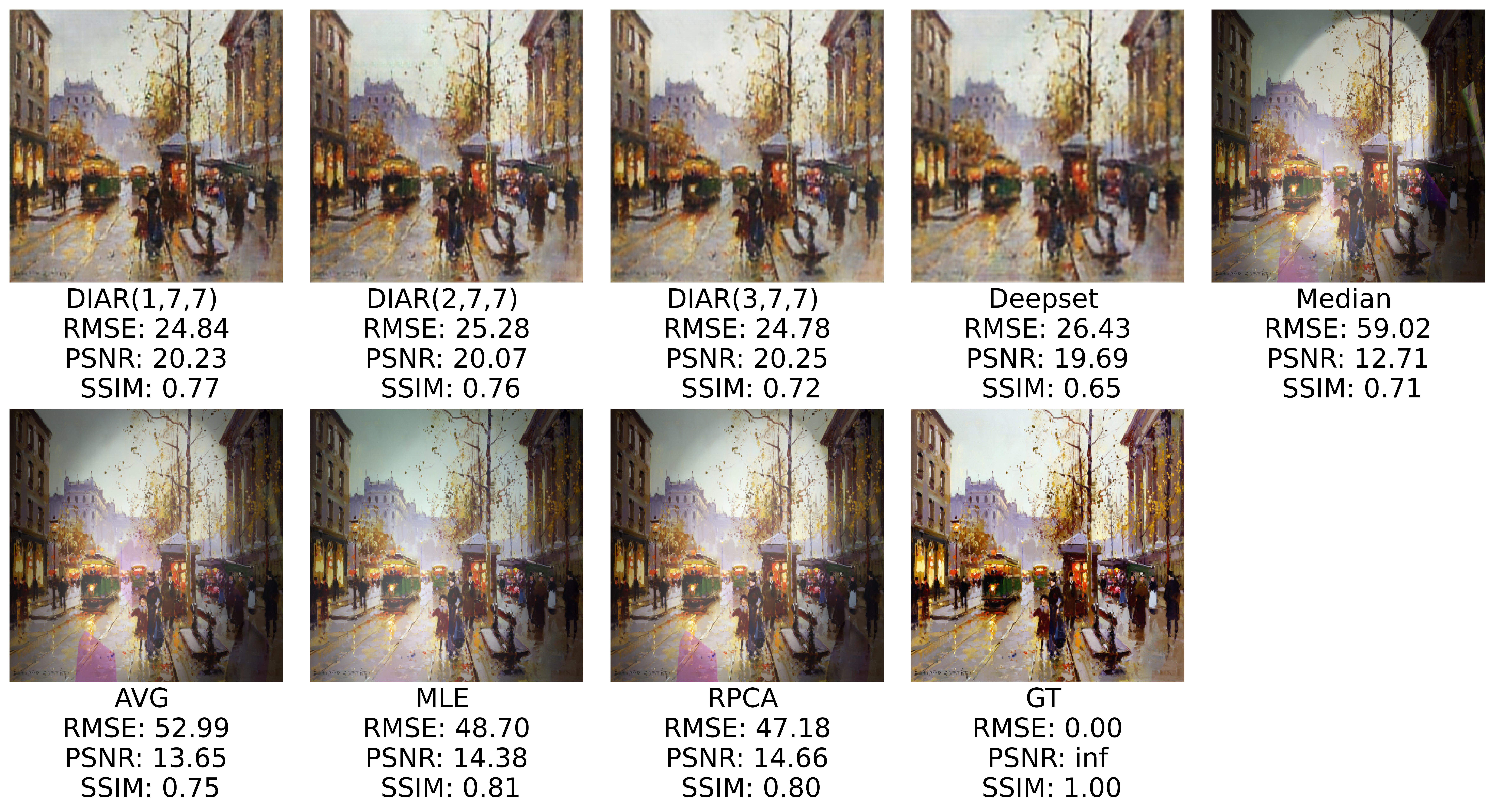}
    \caption{The first row shows a distorted sequence. The results below show the reconstruction of various methods.}
    \label{fig:example1}
\end{figure}

\begin{figure}[H]
    \centering
    \includegraphics[width=\linewidth]{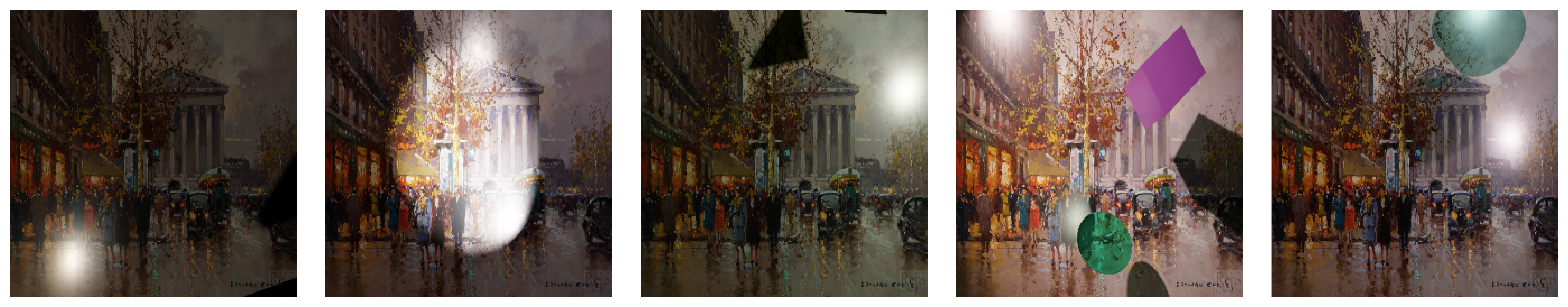}
    \includegraphics[width=\linewidth]{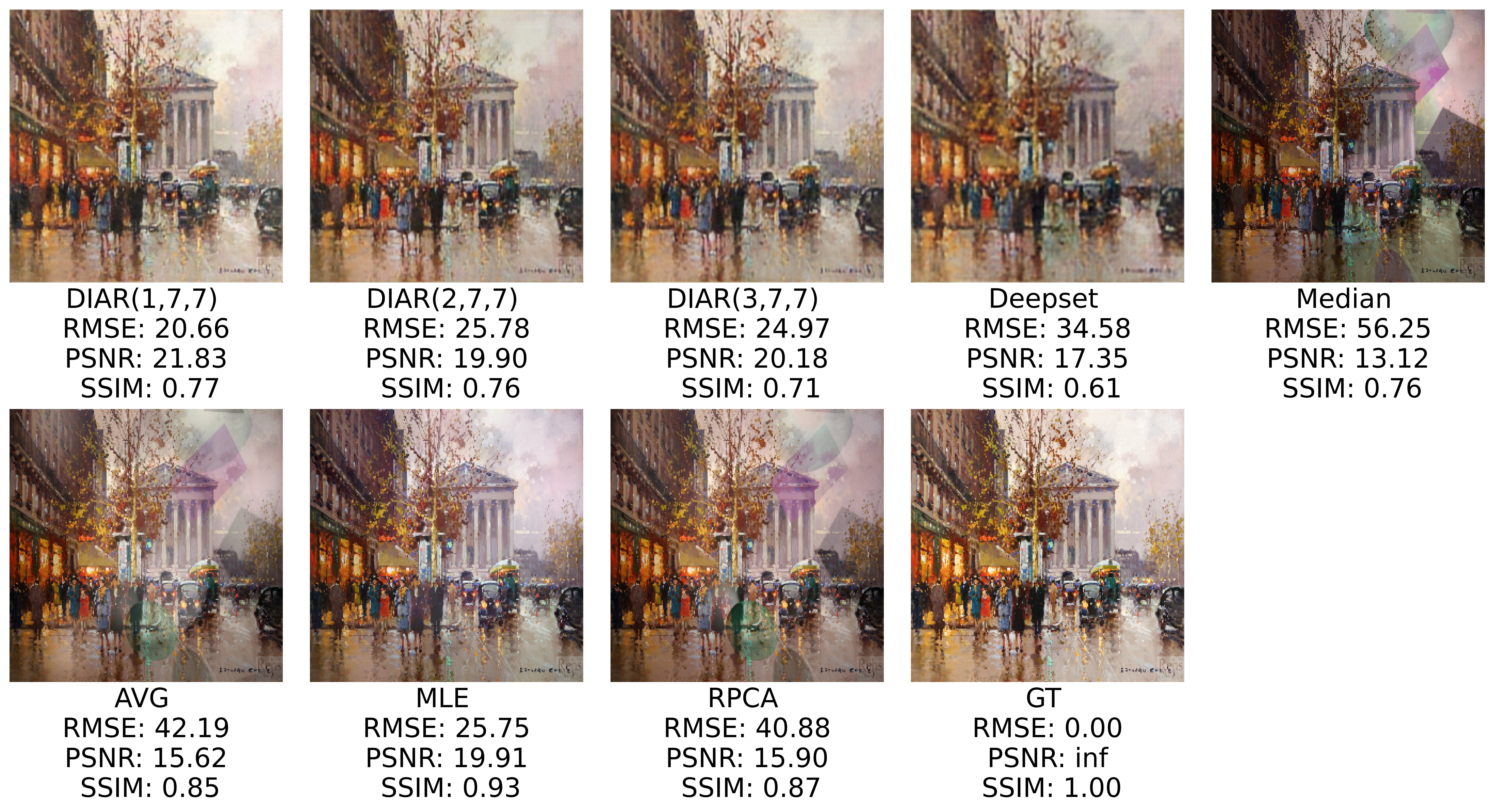}
    \caption{The first row shows a distorted sequence. The results below show the reconstruction of various methods.}
    \label{fig:example2}
\end{figure}

\subsection{Alignment \& Reconstruction:}

Finally, we attempt to align a sequence of distorted images and reconstruct them using an image processing pipeline that consists of differentiable and trainable components. First, we use a pre-trained ResNet to compute dense feature maps for each image. Then we compute pairwise matches using cosine similarity. With the computed matches, we can estimate a homography between each image and a reference frame. We align each image with the reference frame. Finally, we apply our trained neural networks for reconstruction. We also apply our other methods to have a comparison. \\
We measure the quality of the alignment by using our image metrics on the reconstructions. Additionally, we can compare the quality of the homography directly.
We use two error metrics to evaluate the alignment error. 
\subsubsection{1.} 
\begin{figure}[H]
    \centering
    \includegraphics[width=0.45\linewidth]{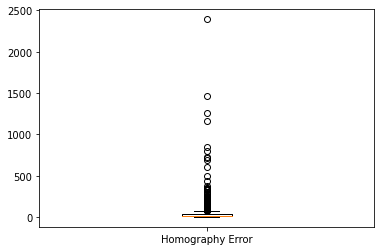}    \includegraphics[width=0.45\linewidth]{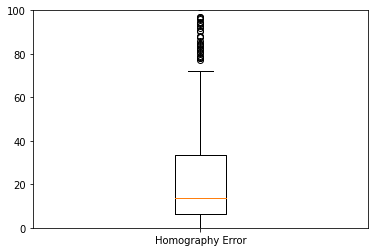}
    \includegraphics[width=0.45\linewidth]{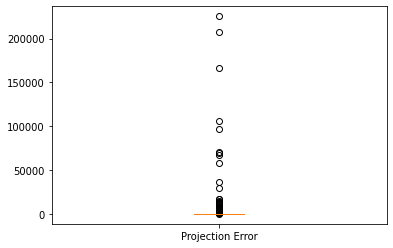}    \includegraphics[width=0.45\linewidth]{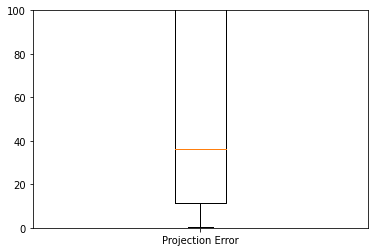}
    \caption{The box plots illustrate the distribution of projection errors and homography errors. The box plots on the right are zoomed in order to see the median. The box plot indicates that the data is skewed and contains large outliers. }
    \label{fig:homog_error}
\end{figure}

Given the ground-truth matrix $H$ and an estimate $H'$, we can't directly take the norm between them. Homographies are equivalent under scale , however $|H-H'| \neq |\lambda H-H'|$. We normalize both homographies, such that their determinants are equal to 1: $det(\hat{H})=det(\hat{H}')=1$. Figure \ref{fig:homog_error} visualizes the distribution of the error over all images.
\subsubsection{2} Given the ground-truth matrix $H$ and an estimate $H'$, we calculate the projection error. We define four fixed point $x_1=(-1,-1),x_2=(1,-1),x_3=(-1,1),x_4=(1,1)$. We measure the average projection error   $\vert Hx_i - H'x_i\vert$

\begin{figure}
    \centering
    \includegraphics[width=0.8\linewidth]{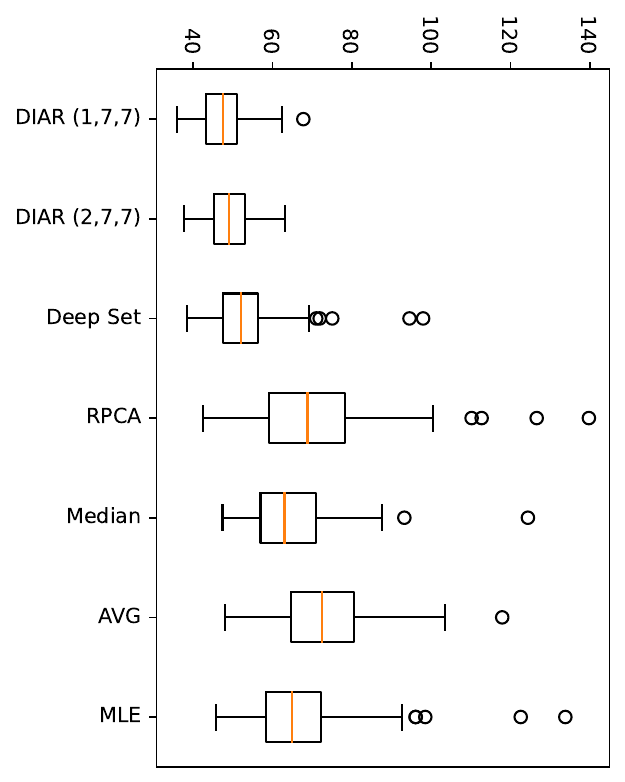}
    \caption{Box plot for RMSE for each reconstruction method}
    \label{fig:rmse_box}
\end{figure}

\begin{figure}
    \centering
    \includegraphics[width=0.8\linewidth]{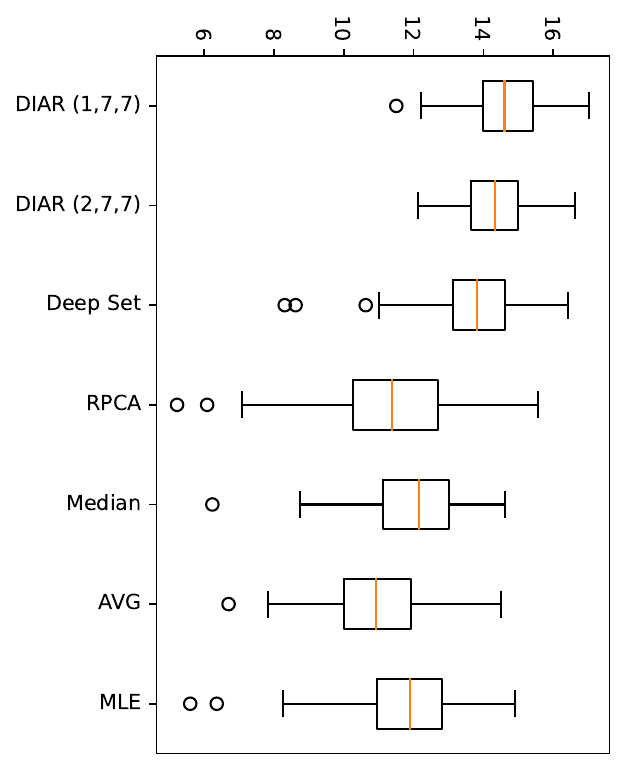}
    \caption{Box plot for PSNR for each reconstruction method}
    \label{fig:psnr_box}
\end{figure}

\begin{figure}
    \centering
    \includegraphics[width=0.8\linewidth]{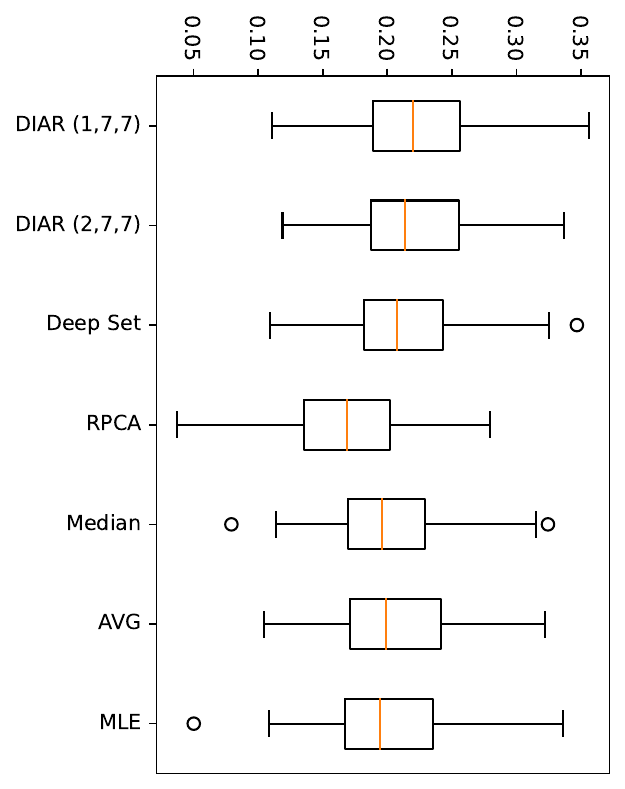}
    \caption{Box plot for SSIM for each reconstruction method}
    \label{fig:ssim_box}
\end{figure}

The boxplots \ref{fig:rmse_box}, \ref{fig:psnr_box} and \ref{fig:ssim_box} show how each reconstruction method deals with the aligned/misaligned images. The transformer seems to be able to handle the perspective distortions better. This might be due to the fact that the Swin transformers learn to understand outliers. They might be better at dealing with complete misalignment or gross distortions. 
\begin{align}
    \hat{H} = \frac{1}{\sqrt[3]{det(H)}} H
\end{align}
 We then compute the norm between them. 

\begin{figure}
    \centering
    \includegraphics[width=1\linewidth]{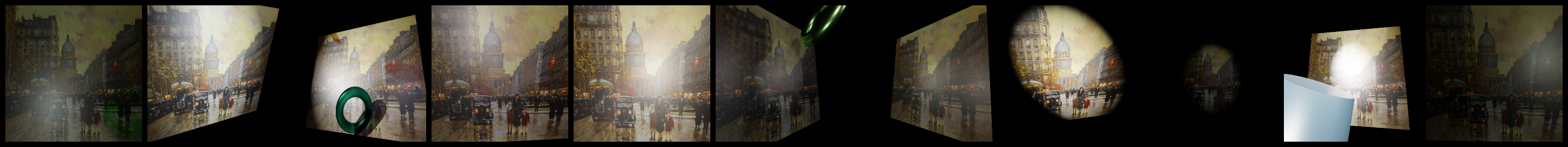}
    \includegraphics[width=1\linewidth]{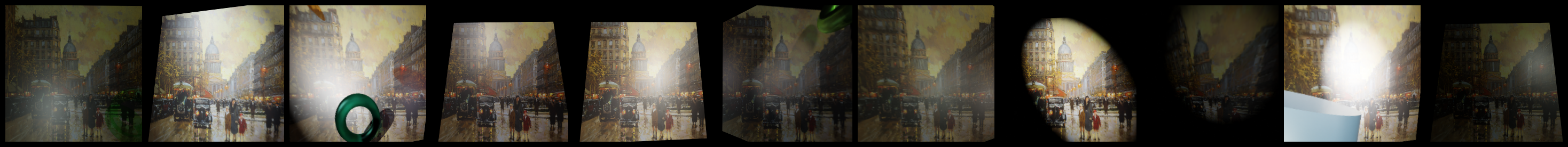}
    \includegraphics[width=1\linewidth]{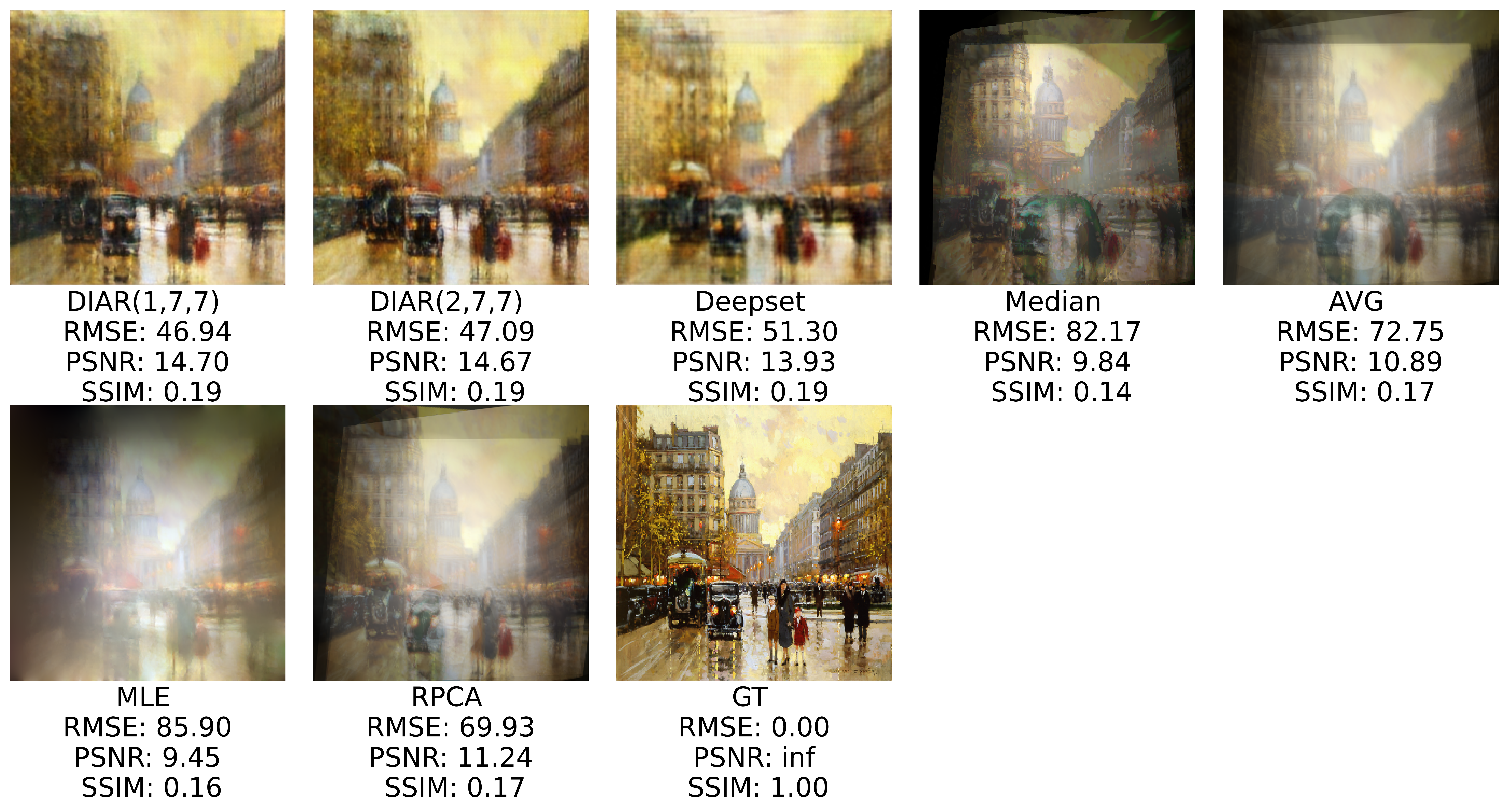}
    \caption{The first row shows a sequence of distorted images. The second row shows the estimated alignment. The results of each reconstruction method are also shown.}
    \label{fig:example_misalign}
\end{figure}

\section{Conclusion}
In this paper, we attempted to solve two problems simultaneously: alignment of images using deep image features and reconstructing the content from distorted images. For this, we created a synthetic dataset that contains various distortions due to lighting, shadows, and occlusion. Furthermore, we added perspective distortions with corresponding ground-truth homographies. We believe that this dataset can be particularly useful for developing robust image descriptors and matching methods due to the variety of distortions and challenging illumination. \\
We implement a general-purpose method for computing point correspondences between images using neural feature maps. Similar methods have been also used in structure-from-motion applications \cite{shen2020ransac,lindenberger2021pixel}. The evaluations show that many of the images are aligned, but there also are strong outliers. The example in figure \ref{fig:example_misalign} shows that the images are aligned. However, the alignment is too coarse for an accurate pixel-wise reconstruction. Further refinement has to be made. A possible improvement is to use bundle adjustment on the computed image matches to improve the initial estimation.
\\
Additionally, we discussed the use of transformers and attention for image aggregation tasks. We used Swin transformers to analyze the temporal dimension more efficiently. This allowed us to improve on the original Deep Set architecture\cite{kwiatkowski2022specularity}. The evaluations showed that transformers enable us to efficiently combine information from multiple images while simultaneously avoiding outliers and artifacts.

%
%
%
%
\newpage
\bibliographystyle{splncs04}
\bibliography{references}

\begin{thebibliography}{10}
\providecommand{\url}[1]{\texttt{#1}}
\providecommand{\urlprefix}{URL }
\providecommand{\doi}[1]{https://doi.org/#1}

\bibitem{arnab2021vivit}
Arnab, A., Dehghani, M., Heigold, G., Sun, C., Lu{\v{c}}i{\'c}, M., Schmid, C.:
  Vivit: A video vision transformer. In: Proceedings of the IEEE/CVF
  International Conference on Computer Vision. pp. 6836--6846 (2021)

\bibitem{hpatches_2017_cvpr}
Balntas, V., Lenc, K., Vedaldi, A., Mikolajczyk, K.: Hpatches: A benchmark and
  evaluation of handcrafted and learned local descriptors. In: CVPR (2017)

\bibitem{bouwmans2018applications}
Bouwmans, T., Javed, S., Zhang, H., Lin, Z., Otazo, R.: On the applications of
  robust pca in image and video processing. Proceedings of the IEEE
  \textbf{106}(8),  1427--1457 (2018)

\bibitem{danielczuk2019segmenting}
Danielczuk, M., Matl, M., Gupta, S., Li, A., Lee, A., Mahler, J., Goldberg, K.:
  Segmenting unknown 3d objects from real depth images using mask r-cnn trained
  on synthetic data. In: Proc. IEEE Int. Conf. Robotics and Automation (ICRA)
  (2019)

\bibitem{detone2016deep}
DeTone, D., Malisiewicz, T., Rabinovich, A.: Deep image homography estimation.
  arXiv preprint arXiv:1606.03798  (2016)

\bibitem{devlin2018bert}
Devlin, J., Chang, M.W., Lee, K., Toutanova, K.: Bert: Pre-training of deep
  bidirectional transformers for language understanding. arXiv preprint
  arXiv:1810.04805  (2018)

\bibitem{dosovitskiy2020image}
Dosovitskiy, A., Beyer, L., Kolesnikov, A., Weissenborn, D., Zhai, X.,
  Unterthiner, T., Dehghani, M., Minderer, M., Heigold, G., Gelly, S., et~al.:
  An image is worth 16x16 words: Transformers for image recognition at scale.
  arXiv preprint arXiv:2010.11929  (2020)

\bibitem{hartley2003multiple}
Hartley, R., Zisserman, A.: Multiple view geometry in computer vision.
  Cambridge university press (2003)

\bibitem{kwiatkowski2022specularity}
Kwiatkowski, M., Hellwich, O.: Specularity, shadow, and occlusion removal from
  image sequences using deep residual sets. In: VISIGRAPP (4: VISAPP). pp.
  118--125 (2022)

\bibitem{lin2021barf}
Lin, C.H., Ma, W.C., Torralba, A., Lucey, S.: Barf: Bundle-adjusting neural
  radiance fields. In: Proceedings of the IEEE/CVF International Conference on
  Computer Vision. pp. 5741--5751 (2021)

\bibitem{lindenberger2021pixel}
Lindenberger, P., Sarlin, P.E., Larsson, V., Pollefeys, M.: Pixel-perfect
  structure-from-motion with featuremetric refinement. In: Proceedings of the
  IEEE/CVF International Conference on Computer Vision. pp. 5987--5997 (2021)

\bibitem{liu2021swin}
Liu, Z., Lin, Y., Cao, Y., Hu, H., Wei, Y., Zhang, Z., Lin, S., Guo, B.: Swin
  transformer: Hierarchical vision transformer using shifted windows. In:
  Proceedings of the IEEE/CVF International Conference on Computer Vision. pp.
  10012--10022 (2021)

\bibitem{liu2022video}
Liu, Z., Ning, J., Cao, Y., Wei, Y., Zhang, Z., Lin, S., Hu, H.: Video swin
  transformer. In: Proceedings of the IEEE/CVF Conference on Computer Vision
  and Pattern Recognition. pp. 3202--3211 (2022)

\bibitem{lowe2004distinctive}
Lowe, D.G.: Distinctive image features from scale-invariant keypoints.
  International journal of computer vision  \textbf{60}(2),  91--110 (2004)

\bibitem{nie2021unsupervised}
Nie, L., Lin, C., Liao, K., Liu, S., Zhao, Y.: Unsupervised deep image
  stitching: Reconstructing stitched features to images. IEEE Transactions on
  Image Processing  \textbf{30},  6184--6197 (2021)

\bibitem{sarlin2020superglue}
Sarlin, P.E., DeTone, D., Malisiewicz, T., Rabinovich, A.: Superglue: Learning
  feature matching with graph neural networks. In: Proceedings of the IEEE/CVF
  conference on computer vision and pattern recognition. pp. 4938--4947 (2020)

\bibitem{shen2020ransac}
Shen, X., Darmon, F., Efros, A.A., Aubry, M.: Ransac-flow: generic two-stage
  image alignment. In: European Conference on Computer Vision. pp. 618--637.
  Springer (2020)

\bibitem{Sun_2021_CVPR}
Sun, J., Shen, Z., Wang, Y., Bao, H., Zhou, X.: Loftr: Detector-free local
  feature matching with transformers. In: Proceedings of the IEEE/CVF
  Conference on Computer Vision and Pattern Recognition (CVPR). pp. 8922--8931
  (June 2021)

\bibitem{vaswani2017attention}
Vaswani, A., Shazeer, N., Parmar, N., Uszkoreit, J., Jones, L., Gomez, A.N.,
  Kaiser, {\L}., Polosukhin, I.: Attention is all you need. Advances in neural
  information processing systems  \textbf{30} (2017)

\bibitem{weiss2001deriving}
Weiss, Y.: Deriving intrinsic images from image sequences. In: Proceedings
  Eighth IEEE International Conference on Computer Vision. ICCV 2001. vol.~2,
  pp. 68--75. IEEE (2001)

\bibitem{zaheer2017deep}
Zaheer, M., Kottur, S., Ravanbakhsh, S., Poczos, B., Salakhutdinov, R.R.,
  Smola, A.J.: Deep sets. Advances in neural information processing systems
  \textbf{30} (2017)

\end{thebibliography}

\end{document}